\pdfoutput=1
\documentclass[11pt,a4paper]{article}
\usepackage[hyperindex,breaklinks]{hyperref}
\usepackage{naaclhlt2019}
\usepackage{times}
\usepackage{latexsym}
\usepackage{tipa}
\usepackage{amsfonts}
\usepackage{amsmath}
\usepackage{breakcites}
\usepackage[moderate]{savetrees}

\usepackage{multirow}
\usepackage{adjustbox}
\usepackage{tabularx}

\usepackage{tikz}
\usepackage[colorinlistoftodos]{todonotes}

\usepackage{url}

\usepackage{booktabs}
\newcommand{\ra}[1]{\renewcommand{\arraystretch}{#1}}

\aclfinalcopy 

\title{Spoken dialect identification in Twitter using a multi-filter architecture}

\author{Mohammadreza Banaei \And
  R\'emi Lebret \\
  EPFL, Switzerland
  \And
  Karl Aberer \\
}

\date{}
\begin{document}
\maketitle

\begin{abstract}
  This paper presents our approach for SwissText \& KONVENS 2020 shared task 2, which is a multi-stage neural model for Swiss German (GSW) identification on Twitter. Our model outputs either GSW or non-GSW and is not meant to be used as a generic language identifier. Our architecture consists of two independent filters where the first one favors recall, and the second one filter favors precision (both towards GSW). Moreover, we do not use binary models (GSW vs. not-GSW) in our filters but rather a multi-class classifier with GSW being one of the possible labels. Our model reaches F1-score of 0.982 on the test set of the shared task.
\end{abstract}
\section{Introduction}
\label{sec:intro}
Out of over 8000 languages in the world~\cite{Hammarstrom:2020}, Twitter language identifier (LID) only supports around 30 of the most used languages\footnote{\url{https://dev.twitter.com/docs/developer-utilities/supported-languages/api-reference}}, which is not enough for NLP community needs. Furthermore, it has been shown that even for these frequently used languages, Twitter LID is not highly accurate, especially when the tweet is relatively short~\cite{zubiaga2016tweetlid}.

However, Twitter data is linguistically diverse and especially includes tweets in many low-resource languages/dialects. Having a better performing Twitter LID can help us to gather large amounts of (unlabeled) text in these low-resource languages that can be used to enrich models in many down-stream NLP tasks, such as sentiment analysis~\cite{volkova-etal-2013-exploring} and named entity recognition
~\cite{ritter2011named}.

However, the generalization of state-of-the-art NLP models to low-resource languages is generally hard due to the lack of corpora with good coverage in these languages. The extreme case is the spoken dialects, where there might be no standard spelling at all. In this paper, we especially focus on Swiss German as our low-resource dialect. As Swiss German is a spoken dialect, people might spell a certain word differently, and even a single author might use different spelling for a word between two sentences. There also exists a dialect continuum across the German-speaking part of Switzerland, which makes NLP for Swiss German even more challenging. Swiss German has its own pronunciation, grammar and also lots of its words are different from German.

There exists some previous efforts for discriminating similar languages with the help of tweets metadata such as geo-location~\cite{williams2017twitter}, but in this paper, we do not use tweets metadata and restrict our model to only use tweet content. Therefore, this model can also be used for language identification in sources other than Twitter.

LIDs that support GSW like fastText~\cite{joulin2016bag} LID model are often trained by using Alemannic Wikipedia, which also contains other German dialects such as Swabian, Walser German, and Alsatian German; hence, these models are not able to discriminate dialects that are close to GSW. Moreover, fastText LID also has a pretty low recall (0.362) for Swiss German tweets, as it identified many of them as German.

In this paper, we use two independently trained filters to remove non-GSW tweets. The first filter is a classifier that favors recall (towards GSW), and the second one favors precision. The exact same idea can be extended to N consecutive filters (with $N\geq2$), with the first $N-1$ favoring recall and the last filter favoring precision. In this way, we make sure that GSW samples are not filtered out (with high probability) in the first $N-1$ iterations, and the whole pipeline GSW precision can be improved by having a filter that favors precision at the end ($N$-th filter). The reason that we use only two filters is that adding more filters improved the performance (measured by GSW F1-score) negligibly on our validation set.

We demonstrate that by using this architecture, we can achieve F1-score of 0.982 on the test set, even with a small amount of available data in the target domain (Twitter data). Section~\ref{sec:model} presents the architecture of each of our filters and the rationale behind the chosen training data for each of them. In section~\ref{sec:experiment}, we discuss our LID implementation details and also discuss the detailed description of used datasets. Section~\ref{sec:result} presents the performance of our filters on the held-out test dataset. Moreover, we demonstrate the contribution of each of the filters on removing non-GSW filters to see their individual importance in the whole pipeline (for this specific test dataset).

\section{Multi-filter language identification}
\label{sec:model}
In this paper, we follow the combination of $N-1$ filters favoring recall, followed by a final filter that favors more precision. We choose $N=2$ in this paper to demonstrate the effectiveness of the approach. As discussed before, adding more filters improved the performance of the pipeline negligibly for this specific dataset. However, for more challenging datasets, it might be needed to have $N>2$ to improve the LID precision.

Both of our filters are multi-class classifiers with GSW being one of the possible labels. We found it empirically better to use roughly balanced classes for training the multi-class classifier, rather than making the same training data a highly imbalanced GSW vs. non-GSW training data for a binary classifier, especially for the first filter (section~\ref{sec:bertfilter}) which has much more parameters compared to the second filter (section~\ref{sec:fasttextfilter}).

\subsection{First filter: fine-tuned BERT model}
\label{sec:bertfilter}

The first filter should be designed in a way to favor GSW recall, either by tuning inference thresholds or by using training data that implicitly enforces this bias towards GSW. Here we follow the second approach for this filter by using different domains for training different labels, which is further discussed below. Moreover, we use a more complex (in terms of the number of parameters) model for the first filter, so that it does the main job of removing non-GSW inputs while having reasonable GSW precision (further detail in section~\ref{sec:result}). The second filter will be later used to improve the pipeline precision by removing a relatively smaller number of non-GSW tweets.

Our first filter is a fine-tuned BERT~\cite{devlin2018bert} model for the LID downstream task. As we do not have a large amount of unsupervised GSW data, it will be hard to train the BERT language model (LM) from scratch on GSW itself. Hence, we use the German pre-trained LM (BERT-base-cased model\footnote{Training details available at {\url{https://huggingface.co/bert-base-german-cased}} \citep{wolf2019transformers}}), which is the closest high-resource language to GSW.

However, this LM has been trained using sentences (e.g., German Wikipedia) that are quite different from the Twitter domain. Moreover, lack of standard spelling in GSW introduces many new words (unseen in German LM training data) that their respective subwords embedding should be updated in order to improve the downstream task performance. In addition, there are even syntactic differences between German and GSW (and even among different variations of GSW in different regions~\cite{honnet2017machine}). For these three reasons, we can conclude that freezing the BERT body (and just training the classifier layer) might not be optimal for this transfer learning between German and our target language. Hence, we also let the whole BERT body be trained during the downstream task, which of course needs a large amount of supervised data to avoid quick overfitting in the fine-tuning phase.

For this filter, we choose the same eight classes for training LID as \newcite{linder2019automatic} (the dataset classes and their respective sizes can be found in section~\ref{sec:dataset}). These languages are similar in structure to GSW (such as German, Dutch, etc.), and we try to train a model that can distinguish GSW from similar languages to decrease GSW false positives. For all classes except GSW, we use sentences (mostly Wikipedia and Newscrawl) from Leipzig text corpora~\citep{goldhahn2012building}. We also use the SwissCrawl~\citep{linder2019automatic} dataset for GSW sentences.

Most GSW training samples (SwissCrawl data) come from forums and social media, which are less formal (in structure and also used phrases) than other (non-GSW) classes samples (mostly from Wikipedia and NewsCrawl). Moreover, as our target dataset consist of tweets (mostly informal sentences), this could make this filter having high GSW recall during the inference phase. Additionally, our main reason for using a cased tokenizer for this filter is to let the model also use irregularities in writing, such as improper capitalization. As these irregularities mostly occur in informal writing, it will again bias the model towards GSW (improving GSW recall) when tweets are passed to it, as most of the GSW training samples are informal.

\subsection{Second filter: fastText classifier}
\label{sec:fasttextfilter}
For this filter, we also train a multiclass classifier with GSW being one of the labels. The other classes are again close languages (in structure) to GSW such as German, Dutch and Spanish (further detail in section~\ref{sec:dataset}). Additionally, as mentioned before, our second filter should have a reasonably high precision to enhance the full pipeline precision. Hence, unlike the first filter, we choose the whole training data to be sampled from a similar domain to the target test set. non-GSW samples are tweets from SEPLN 2014~\citep{zubiaga2014overview} and \newcite{carter2013microblog} dataset. GSW samples consist of this shared task provided GSW tweets and also part of GSW samples of Swiss SMS corpus~\citep{sms4science} dataset.

As the described training data is rather small compared to the first filter training, we should also train a simpler architecture with significantly fewer parameters. We take advantage of fastText~\citep{joulin2016bag} for training this model, which is based on a bag of character $n$-grams in our case. Moreover, unlike the first filter, this model is not a cased model, and we make input sentences lower-case to reduce vocab size. Our used hyper-parameters for this model can be found in section~\ref{sec:experiment}.

\section{Experimental Setup}
\label{sec:experiment}
In this section, we describe the datasets and the hyper-parameters for both filters in the pipeline. We also describe our preprocessing method that is specifically designed to handle inputs from social media.

\subsection{Datasets}
For both filters, we use 80\% of data for training, 5\% for validation set and 15\% for the test set.
\label{sec:dataset}
\subsubsection{First filter}
The sentences are from Leipzig corpora~\citep{goldhahn2012building} and SwissCrawl~\citep{linder2019automatic} dataset. The classes and the number of samples in each class are shown in Table \ref{tab:filter1_dataset}. We pick the proposed classes by \newcite{linder2019automatic} for training GSW LID. The main differences of our first filter with their LID are the GSW sentences and the fact that our fine-tuning dataset is about three times larger than theirs. Each of ``other''\footnote{Catalan, Croatian, Danish, Esperanto, Estonian, Finnish, French, Irish, Galician, Icelandic, Italian, Javanese, Konkani, Papiamento, Portuguese, Romanian, Slovenian, Spanish, Swahili, Swedish} and ``GSW-like''\footnote{Bavarian, Kolsch, Limburgan, Low German, Northern Frisian, Palatine German} classes are a group of languages where their respective members cannot be represented as a separate class due to having a small number of samples. The GSW-like is included to make sure that the model can distinguish other German dialects from GSW (hence, reducing GSW false positives).

\begin{table}[h]
    \centering	
    \tabcolsep=0.14cm
	\ra{0.9}
	\begin{tabular}{lc}
	    \toprule
	    Language     & Number of samples \\ 
		\midrule
        Afrikaans       & 250000  \\ 
        German       & 100000  \\ 
        English       & 250000  \\ 
        Swiss-German       & 250000  \\ 
        GSW-like       & 250000  \\ 
        Luxembourgian       & 250000  \\ 
        Dutch       & 250000  \\  
        Other       & 250000 \\ \bottomrule
    \end{tabular}
    \caption{Distribution of samples in the first filter dataset}
    \label{tab:filter1_dataset}
\end{table}
\subsubsection{Second filter}
The sentences are mostly from Twitter (except for some GSW samples from Swiss SMS corpus~\citep{sms4science}). In Table \ref{tab:filter2_dataset}, we can see the distribution of different classes. The GSW samples consist of 1971 tweets (provided by shared task organizers) and 3000 GSW samples from Swiss SMS corpus.
\begin{table}[h]
    \centering	
    \tabcolsep=0.14cm
	\ra{0.9}
	\begin{tabular}{lc}
	    \toprule
	    Language     & Number of samples \\ 
		\midrule
        Catalan       & 2000  \\ 
        Dutch       & 560  \\ 
        English       & 2533  \\ 
        French       & 639  \\ 
        German       & 3608  \\ 
        Spanish       & 2707  \\ 
        Swiss German       & 4971  \\ \bottomrule
    \end{tabular}
    \caption{Distribution of samples in the second filter dataset}
    \label{tab:filter2_dataset}
\end{table}

\subsection{Preprocessing}

As the dataset sentences are mostly from social media, we used a custom tokenizer that removes common social media tokens (emoticons, emojis, URL, hashtag, Twitter mention) that are not useful for LID. We also normalize word elongation as it might be misleading for LID. In the second filter, we also make the input sentences lower-case before passing it to the model.

\subsection{Implementation details}
\label{sec:implementation_details}
\subsubsection{BERT filter}
We train this filter by fine-tuning a German pre-trained BERT-cased model on our LID task. As mentioned before, we do not freeze the BERT body in the fine-tuning phase. We train it for two epochs, with a batch size of 64 and max-seq-length of 64. We use Adam optimizer~\citep{kingma2014adam} with a learning rate of 2e-5.

\subsubsection{fastText filter}
We train this filter using fastText~\citep{joulin2016bag} classifier for 30 epochs using character $n$-grams as features (where $2\leq n \leq 5$) and the embedding dimension set to 50. To favor precision during inference, we label a tweet as GSW if the model probability for GSW is greater than 64\% (this threshold is seen as a hyper-parameter and was optimized according to validation set).

\section{Results}
\label{sec:result}

In this section, we evaluate our two filters performance (either in isolation or when present in the full pipeline) on the held-out test dataset of the shared task. We also evaluate the BERT filter on its test data (Leipzig and SwissCrawl samples).

\subsection{BERT filter performance on Leipzig + SwissCrawl corpora}
We first evaluate our BERT filter on the test set of the first filter (Leipzig corpora + SwissCrawl). In Table \ref{tab:FFevaluationleipzig} we demonstrate the filter performance on different labels. The filter has an F1-score of 99.8\% on the GSW test set. However, when this model is applied to Twitter data, we expect a decrease in performance due to having short and also informal messages.

\begin{table}[h]
    \centering	
    \tabcolsep=0.14cm
	\ra{0.9}
	\begin{tabular}{lccc}
	    \toprule
	    Language     & Precision & Recall & F1-score \\ 
		\midrule
        Afrikaans       & 0.9982 & 0.9981 & 0.9982  \\ 
        German       & 0.9976 & 0.9949 & 0.9962 \\ 
        English       & 0.9994 & 0.9992 & 0.9993 \\ 
        Swiss-German       & 0.9974 & 0.9994 & 0.9984 \\ 
        GSW-like       & 0.9968 & 0.9950 & 0.9959 \\ 
        Luxembourgian       & 0.9994 & 0.9989 & 0.9992  \\ 
        Dutch       & 0.9956 & 0.9965 & 0.9960\\  
        Other       & 0.9983 & 0.9989 & 0.9986\\ \bottomrule
    \end{tabular}
    \caption{First filter performance on Leipzig + SwissCrawl corpora}
    \label{tab:FFevaluationleipzig}
\end{table}

\subsection{Performance on the shared-task test set}
In Table~\ref{tab:finaltestset}, we can see both filters performance either in isolation or when they are used together. As shown in this table, the model improvement by adding the second filter is rather small. The main reason can be seen in Table~\ref{tab:non_gsw_removal} as the majority of non-GSW filtering is done by the first filter for the shared-task test set (Table~\ref{tab:test_dataset_dist}).

\begin{table}[h]
    \centering	
    \tabcolsep=0.14cm
	\ra{0.9}
	\begin{tabular}{lccc}
	    \toprule
	    Model     & Precision & Recall & F1-score \\ 
		\midrule
        BERT filter       & 0.9742 & 0.9896 & 0.9817  \\ 
        fastText Filter       & 0.9076 & 0.9892 & 0.9466 \\ 
        BERT + fastText       & 0.9811 & 0.9834 & 0.9823 \\  
        fastText Baseline       & 0.9915 & 0.3619 & 0.5303 \\ \bottomrule
    \end{tabular}
    \caption{Filters performance on the shared task test-set compared to fastText~\citep{joulin2016bag} LID baseline}
    \label{tab:finaltestset}
\end{table}

\begin{table}[h]
    \centering	
    \tabcolsep=0.14cm
	\ra{0.9}
	\begin{tabular}{lc}
	    \toprule
	    Model     & Number of filtered samples \\ 
		\midrule
        BERT filter       & 2741  \\ 
        fastText Filter       & 35 \\  \bottomrule
    \end{tabular}
    \caption{Number of Non-GSW removals by each filter}
    \label{tab:non_gsw_removal}
\end{table}

\begin{table}[h]
    \centering	
    \tabcolsep=0.14cm
	\ra{0.9}
	\begin{tabular}{lc}
	    \toprule
	    Label     & Number of samples \\ 
		\midrule
        not-GSW      & 2782  \\ 
        GSW       & 2592 \\  \bottomrule
        
    \end{tabular}
    \caption{Distribution of labels in test set}
    \label{tab:test_dataset_dist}
\end{table}

\subsection{Discussion}
Our designed LID outperforms the baseline significantly (Table~\ref{tab:finaltestset}) which underlines the importance of having a domain-specific LID. Additionally, although the positive effect of the second filter is quite small on the test set, when we applied the same architecture on randomly sampled tweets (German tweets according to Twitter API), we observed that having the second filter could reduce the number of GSW false positives significantly. Hence, the number of used filters is indeed totally dependent on the complexity of the target dataset.

\section{Conclusion }
In this work, we propose an architecture for spoken dialect (Swiss German) identification by introducing a multi-filter architecture that is able to filter out non-GSW tweets during the inference phase effectively. We evaluated our model on the GSW LID shared task test-set, and we reached an F1-score of 0.982.

However, there are other useful features that can be used during training, such as orthographic conventions in GSW writing, as observed by \newcite{honnet2017machine}, which their presence might not be easily captured even by a complex model like BERT. Moreover, in this paper, we did not use tweets metadata as a feature and only focused on tweet content, although they can improve LID classification for dialects considerably~\citep{williams2017twitter}. These two, among others, are future works that need to be further studied to see their usefulness for low-resource language identification.

\bibliography{naaclhlt2019}
\bibliographystyle{acl_natbib}

\end{document}